\documentclass[conference, 10pt]{IEEEtran} 
\usepackage{amsmath,amssymb,bm} 
\usepackage{bbm}
\usepackage{graphicx} 
\usepackage[acronym]{glossaries}
\usepackage{tikz}
\usepackage{pgfplots}
\usepackage{algpseudocode, algorithm}
\pgfplotsset{compat=1.18}
\usetikzlibrary{shapes.multipart}

\usetikzlibrary{positioning,arrows.meta,fit}
\usetikzlibrary{shapes,arrows.meta}

\usepackage[caption=false]{subfig}

\newacronym{gogenzip}{GO-GenZip}{Goal-Oriented Generative Sampling and Hybrid Compression}
\newacronym{mae}{MaskAE}{Masked Autoencoder}
\newacronym{ib}{IB}{Information Bottleneck}
\newacronym{go}{GO}{Goal-Oriented}
\newacronym{genai}{GenAI}{Generative AI}
\newacronym{vae}{VAE}{Variational Autoencoder}
\newacronym{st}{ST}{Straight-through}
\newacronym{mec}{MEC}{Mobile Edge Computing}
\newacronym{bs}{BS}{Base station}
\newacronym{kpi}{KPI}{Key Performance Indicator}
\newacronym{dcap}{DCAP}{Data Collection and Analysis Platform}
\newacronym{ml}{ML}{Machine Learning}
\newacronym{ae}{AE}{Autoencoder}
\newacronym{isac}{ISAC}{Integrated Sensing and Communication}
\newacronym{rl}{RL}{Reinforcement Learning}
\newacronym{lzma}{LZMA}{Lempel-Ziv Markov chain algorithm}
\newacronym{sr}{SR}{sampling ratio}
\newacronym{cr}{CR}{compression ratio}
\newacronym{maer}{MAE}{Mean Absolute Error}

\newcommand{\mc}[1]{\mathcal{#1}}   
\newcommand{\mb}[1]{\mathbf{#1}}    
\newcommand{\ve}[1]{\boldsymbol{\mathbf{#1}}} 
\newcommand{\field}[1]{\mathbb{#1}}
\newcommand{\operator}[1]{\mathrm{#1}}

\newcommand{\EE}{{\field{E}}} 
\newcommand{\R}{{\field{R}}}

\newcommand{\Gs}{{\mc{G}}}

\newcommand{\Ks}{{\mc{K}}}
\newcommand{\Xs}{{\mc{X}}}
\newcommand{\Ys}{{\mc{Y}}}
\newcommand{\Zs}{{\mc{Z}}}
\newcommand{\Sss}{{\mc{S}}}
\newcommand{\Rs}{{\mc{R}}}
\newcommand{\Ns}{{\mc{N}}}

\newcommand{\mX}{{\mb{X}}}
\newcommand{\mY}{{\mb{Y}}}
\newcommand{\mZ}{{\mb{Z}}}

\newcommand{\vm}{\ve{m}}

\newcommand{\vc}{\ve{c}}

\definecolor{mplblue}{HTML}{1f77b4}
\definecolor{mplorange}{HTML}{ff7f0e}
\definecolor{mplgreen}{HTML}{2ca02c}

\IEEEoverridecommandlockouts 
\begin{document}

\title{GO-GenZip: Goal-Oriented Generative Sampling and Hybrid Compression}

\author{\IEEEauthorblockN{Pietro Talli, Qi Liao, Alessandro Lieto, Parijat Bhattacharjee, Federico Chiariotti, and Andrea Zanella\thanks{P. Talli, F. Chiariotti, and A. Zanella (emails: pietro.talli@phd.unipd.it, federico.chiariotti@unipd.it, andrea.zanella@unipd.it) are with the Dept. of Information Engineering, University of Padova, Italy. Q. Liao and A. Lieto (emails: qi.liao@nokia-bell-labs.com, alessandro.lieto@nokia-bell-labs.com) are with Nokia Bell Labs Stuttgart, Germany. P. Bhattacharjee (email: parijat.bhattacharjee@nokia.com) is with Nokia Bengaluru, India. This work was in part supported by the European Union's NextGenerationEU framework, as part of the Italian National Recovery and Resilience Plan (NRRP), under the RESTART partnership on ``Telecommunications of the Future'' (PE0000001).}}}

\maketitle

\begin{abstract}
Current network data telemetry pipelines consist of massive streams of fine-grained Key Performance Indicators (KPIs) from multiple distributed sources towards central aggregators, making data storage, transmission, and real-time analysis increasingly unsustainable. This work presents a generative AI (GenAI)-driven sampling and hybrid compression framework that redesigns network telemetry from a goal-oriented perspective. Unlike conventional approaches that passively compress fully observed data, our approach jointly optimizes what to observe and how to encode it, guided by the relevance of information to downstream tasks. The framework integrates adaptive sampling policies, using adaptive masking techniques, with generative modeling to identify patterns and preserve critical features across temporal and spatial dimensions. The selectively acquired data are further processed through a hybrid compression scheme that combines traditional lossless coding with GenAI-driven, lossy compression. Experimental results on real network datasets demonstrate over 50$\%$ reductions in sampling and data transfer costs, while maintaining comparable reconstruction accuracy and goal-oriented analytical fidelity in downstream tasks.
\end{abstract}

\begin{tikzpicture}[remember picture, overlay]
      \node[draw,minimum width=3.in] at ([yshift=-1cm]current page.north)  {Accepted for publication at IEEE Internation Conference on Communications  (ICC) 2026.};
\end{tikzpicture}
\vspace{-0.4cm}
\section{Introduction}
Next-generation networks are defined by a growing need for adaptability, driven by diverse services and dynamic operational contexts, which are critically dependent on robust monitoring and telemetry data \cite{wen2022fine}. However, the exponential growth of telemetry data, which includes traffic patterns, channel state information, user attributes, and mobility profiles, places significant strain on network memory, bandwidth, storage, and processing resources. Traditional, centralized monitoring paradigms that aggregate raw telemetry from distributed terminals are increasingly inadequate to manage the complexity introduced by \gls{mec} and disaggregated architectures. Consequently, efficient collection, processing, and distributed analysis of telemetry data are key enablers of self-organization capabilities~\cite{celdran2017automatic}.

In parallel, network management is progressively shifting towards modern \gls{ml}-based solutions, which pose a new challenge to system telemetry: identifying and acquiring the most relevant data for each learning objective. In practice, excessive data collection remains common due to the computational complexity of relevance estimation methods \cite{jovic2015review} (e.g., Shapley values~\cite{rozemberczki2022shapley}), leading to inefficient telemetry pipelines. 
Recent advances in \gls{genai} and autoencoder-based compression offer a promising path toward adaptive and efficient data reduction~\cite{hinton2006reducing,minnenbt18,cheng2020image}. \glspl{mae} can reconstruct high-dimensional data from partial observations~\cite{he2022masked}, yet they typically rely on random masking strategies that ignore data structure, context, and task objectives. This randomness limits their efficiency when applied to structured, high-dimensional telemetry data, where intelligent selection of observed entries is crucial. A  promising approach in this direction is goal-oriented communication (GO), which focuses on transmitting only relevant information for a task or a goal ~\cite{liao2024adasem, talli2024effective}.

Building on these principles, we propose \gls{gogenzip}, a goal-oriented generative compression framework for network telemetry that adaptively samples and compresses data based on contextual and task information. \gls{gogenzip} integrates \gls{mae}-based generative compression with traditional lossless coding. 
This strategy allows for a dramatic reduction in the collected and transmitted telemetry data, while preserving the performance of the \gls{ml} algorithms that use such data.  

The contribution of this work can be summarized as follows:
\begin{itemize}
    \item We design an adaptive masking policy to sample and monitor a subset of relevant telemetry data to maximize \gls{go} performance. 
    \item We introduce a hybrid compression policy to balance the tradeoff between reconstruction fidelity and compression efficiency by combining lossy compression based on \gls{genai} with classical lossless methods.
    \item We propose a \gls{go} end-to-end training method to jointly optimize masking and compression policies, thus ensuring task-driven data efficiency across multiple objectives.
    \item We validate the proposed framework on real network telemetry data collected from more than 1,000 operational \glspl{bs}, demonstrating significant gains in efficiency and accuracy compared to policies with fixed and generative-only baselines.
\end{itemize}
Our proposed solution is sufficiently general to be applied in diverse contexts and with various types of data. Since we identify the transfer and processing of large tensors as fundamental challenges in future networks, we believe that this system could also prove valuable in other use cases such as channel charting and \gls{isac} scenarios.
\section{System Model and Problem Formulation}\label{sec:ProbState}
\begin{figure}
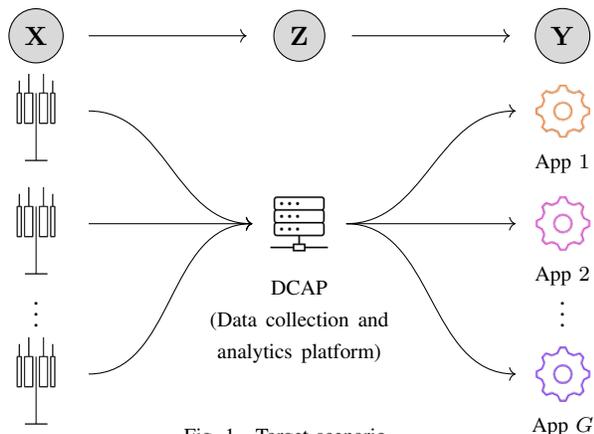

    \centering
    \include{scheme}
    \vspace{-1.5cm}
    \caption{Target scenario.} 
    \label{fig:scheme} 
    \vspace{-.5cm}
\end{figure}

We consider a multi-source, multi-task network telemetry serving a  system of $N$ \glspl{bs}, belonging to a set $\Ns$, and $G$ network applications with goals $\Gs = \{\text{App $1$, ..., App $G$\}}$, as shown in Fig.~\ref{fig:scheme}. Each \gls{bs} collects a set of $K$ \glspl{kpi}, denoted by $\Ks$, periodically sampled every, e.g., $15$ minutes or one  hour. We assume time is slotted and, at each time step $t$, each \gls{bs} $n\in\Ns$ transfers to the \gls{dcap} server its local \gls{kpi} tensor, $\mX_n(t)$. This vector collects the \gls{kpi} samples gathered over a predefined period of time $T_n$, so that we have $\mX_n(t)\in\R^{K\times T_n}$. To generalize the notation across the different \glspl{bs}, which collect and transmit measurements independently and potentially with different periods, hereafter we indicate as $\mX\in\R^D$ the data collected by a generic \gls{bs}, where $D$ is the flattened dimensionality of the measurable data. This makes it possible to formulate the subsequent model and policies in general terms, suitable for different \glspl{bs} and reporting periods.

\subsection{The Classical Goal-Oriented Compression Problem}

Classical goal-oriented compression has been studied mainly through the lens of the \gls{ib} principle \cite{tishby2000information, butakov2023information}, which finds a compressed representation $ \mZ\in\Zs$ of an input $\mX\in\Xs$ that preserves relevant information about a related variable $\mY\in\Ys$, while minimizing the information about $\mX$ itself (the ``bottleneck''). Mathematically,
\begin{equation}
\label{eq:ib}
    \inf_{p(z\mid x)} \; I(\mX; \mZ) - \beta I(\mZ; \mY),
\end{equation}
where $I(A;B)$ denotes the mutual information between $A$ and $B$, and $\beta$ controls the trade-off between removing irrelevant information from $\mX$ and retaining the components that predict $\mY$ through the compressed representation $\mZ$.

This formulation focuses solely on the compression stage, that is, on deciding \emph{what information to transmit}. It is also worth noting that this legacy view treats the source $\mb{X}$ as fully observed and optimizes only the compression stage. In contrast, we are interested in studying the \emph{joint problem of adaptive sampling and hybrid compression}, i.e., not only what to transmit but also \emph{what to observe} in the first place.

\subsection{Goal-Oriented Sampling and Hybrid Compression Problem}

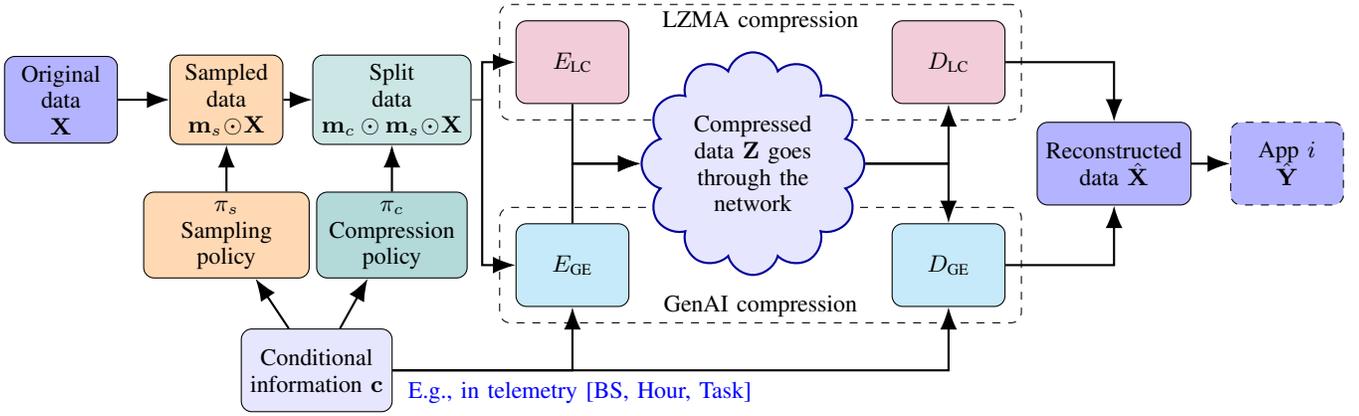
\begin{figure*}
    \centering
    \begin{tikzpicture}[
  node distance=22mm,
  block/.style={rectangle, draw, rounded corners, minimum height=11mm, minimum width=15mm, align=center},
  process/.style={rectangle, draw, fill=orange!30, rounded corners, align=center, minimum width=22mm},
  policy/.style={rectangle, draw, fill=teal!30, rounded corners, align=center, minimum width=20mm},
  dashedbox/.style={draw, dashed, rounded corners, inner sep=6pt},
  arrow/.style={-{Latex[length=3mm]}, thick},
  every node/.style={font=\small}, 
  cloudnode/.style={
    cloud, draw=blue!60!black, thick,
    fill=blue!10,
    cloud puffs=12, cloud puff arc=150,
    minimum width=1.5cm, minimum height=0.2cm,
    align=center
  }
]

\node[block, fill=blue!30] (orig) {Original\\data\\$\mb{X}$};
\node[block, fill=orange!30, right of=orig] (sampled) {Sampled\\data\\$\mb{m}_s\!\odot\!\mb{X}$};
\node[block, fill=teal!20, right of=sampled] (split) {Split\\data\\$\mb{m}_c \odot \mb{m}_s\!\odot\!\mb{X}$};

\node[block, fill=purple!20, right of=split, node distance=24mm, yshift=5mm] (tent) {$E_{\text{LC}}$};
\node[block, fill=purple!20, right of=tent, node distance=50mm] (tinv) {$D_{\text{LC}}$};

\node[block, fill=cyan!20, below of=tent, node distance=20mm, yshift=-7mm] (egen) {$E_{\text{GE}}$};
\node[block, fill=cyan!20, right of=egen, node distance=50mm] (dgen) {$D_{\text{GE}}$};
\node[block, fill=blue!30, right of=dgen, node distance=22mm, yshift=13.5mm] (recon) {Reconstructed\\data $\hat{\mb{X}}$};

\coordinate[right of=split, node distance=12mm](dot) {};

\draw[arrow] (orig) -- (sampled);
\draw[arrow] (sampled) -- (split);
\draw (split) -- (dot);
\draw[arrow] (dot) |- (tent);
\draw[arrow] (tinv) -| (recon);
\draw[arrow] (dot) |- (egen.west);
\draw[arrow] (dgen) -| (recon);

\node[process, below of=sampled, node distance=18mm] (pi_mask) {$\pi_{s}$\\Sampling\\policy};
\node[policy,  below of=split,   node distance=18mm] (pi_comp) {$\pi_{c}$\\Compression\\policy};
\draw[arrow] (pi_mask) -- (sampled);
\draw[arrow] (pi_comp) -- (split);

\node[block, fill=blue!10, below of=pi_mask, node distance=18mm, xshift=12mm] (cond) {Conditional\\information $\mb{c}$};
\draw[arrow] (cond) -- (pi_mask);
\draw[arrow] (cond) -- (pi_comp);
\node[align=left, right of=cond, node distance=3.5cm, yshift=-3mm] (ctx)
  {\textcolor{blue}{E.g., in telemetry [BS, Hour, Task]}};

\node[dashedbox, fit=(tent)(tinv), label={[align=center,yshift=-5mm]above:LZMA compression}] (lzma_box) {};
\node[dashedbox, fit=(egen)(dgen), label={[align=center,yshift=5mm]below:GenAI compression}] (genai_box) {};

\node[cloudnode, right of = tent, yshift=-13.5mm, xshift=2mm] (net) {Compressed \\data $\mb{Z}$ goes\\ through the\\network};

\draw[arrow] (tent) |- (net);
\draw[arrow] (egen) |- (net);
\draw[arrow] (net) -| (tinv);
\draw[arrow] (net) -| (dgen);
\draw[arrow] (cond) -| (egen);


\draw[arrow] (cond) -| (dgen);

\node[block, dashed, fill=blue!30, right of=recon, node distance=11mm, xshift=12mm] (task) {App $i$ \\ $\hat{\mb{Y}}$};

\draw[arrow] (recon) -- (task);

\end{tikzpicture}
    \vspace{-1cm}
    \caption{GO-GenZip sampling and compression architecture.}
     \label{fig:architecture}
     \vspace{-0.4cm}
\end{figure*}

Modern telemetry systems face two main constraints: (1) source sampling, which incurs monitoring and storage costs, and (2) transmission of encoded data, which drives the communication cost. These challenges are amplified for high-dimensional tensor data. To address them, we propose an adaptive encoding function that performs joint sampling and hybrid compression. Here, ``hybrid'' denotes the ability to combine different compression strategies, including \emph{classical lossless compression} and \emph{generative autoencoders}, while a policy learns to adapt to varying contexts. Controlling the fraction of data encoded with lossless compression enables an adaptive management of the bottleneck. Lossless compression introduces no distortion into the original data, thereby preserving mutual information and high fidelity, particularly for sparse data. Conversely, the generative autoencoder compresses data into a small, lossy latent space, so it provides an approximate measurement at a much lower communication cost, as well as allowing the receiver to infer unsampled data. However, training autoencoders to efficiently represent sparse data and preserve reconstruction quality is a complex task, motivating the need for a hybrid approach.

We aim to design sampling and hybrid compression policies that depend only on \emph{low-cost adaptive context information} $\vc\in\R^C$, with $C\ll K$. For the network telemetry use case, in particular, context information can include embeddings of \emph{\gls{bs} class}, \emph{hour index}, and \emph{task index}. This approach makes it possible to adapt the sampling sand compression strategies to the context information, rather than using a single fixed sampling mask and compression scheme for all conditions. The two policies are defined as follows:

\begin{itemize}
\item \emph{sampling policy}: let  $\vm_s\in \{0,1\}^D$ be the binary sampling mask of the observable data space $D$, where element $i$ is collected (or sampled) iff $\vm_s[i] = 1$. The sampled components can therefore be written as $\mb{X}_s = \mb{m}_s \odot \mb{X}$, where $\odot$ represents the element-wise product. The sampling policy can be defined as $\pi_s:\R^C\to \mc{P}\left(\{0, 1\}^D\right)\subseteq[0, 1]^D$, where $\mc{P}$ denotes the set of distributions. Given context information $\vc$, $\pi_s(\vm_s|\vc)$ defines the probability that the sampling mask $\vm_s$ is selected. 
\item \emph{hybrid compression policy}: let $\mathbf{m}_c \in {0,1}^D$ be the ``compression selector", where $\mathbf{m}_c[i]=1$ indicates compression by a generative autoencoder, and $\mathbf{m}_c[i]=0$ by classical lossless coding for each sample entry $i$. The hybrid compression policy is therefore defined as $\pi_c:\R^C\to \mc{P}\left(\{0, 1\}^D\right)\subseteq[0, 1]^D$, such that $\pi_c(\vm_c|\vc)$ is the probability of choosing the compression selector $\vm_c$ given the context $\vc$. 
\end{itemize}
The encoder and decoder use masks $\mZ$ and $\hat{\mY}$, defined as 
\begin{equation}
    \mZ = f(\mX_s, \vm_c), ~ \hat{\mY} = g(\mZ, \vm_c), 
    \label{eq:maskedCodec}
\end{equation}
with $\vm_c\!\sim\!\pi_c(\cdot|\vc)$. The \emph{expected sampling cost} is modeled by
\begin{equation}
    \Sss(\pi_s) \triangleq \EE_{\vc, \vm_s}\left[\sum_{i=1}^D c_i\vm_s[i]\right],
    \label{eq:sampleCost}
\end{equation}
where $\vm_s\sim\pi_s(\cdot|\vc)$ and $c_i$ is the sampling cost per-entry. We often have $c_i= 1$, $\forall i$, so that $\Sss(\cdot)$ is the expected number of sampled entries. The \emph{expected rate} of the hybrid scheme is then defined as
\begin{align}
     \Rs(\pi_s, \pi_c, f)  &=   
    \EE_{\vc,\vm_s, \vm_c}\Bigg[\mathbbm{1}_{\{\vm_s^T\vm_c >0\}}R_{\operator{GE}}(\mX)  \nonumber\\
        & + \frac{1}{D}\sum_{i=1}^D  \vm_s[i] \left(1-\vm_c[i])R_{\operator{LC}}\right)\Bigg],
    \label{eq:rateCost}
\end{align}
where $R_{\operator{GE}}$ and $R_{\operator{LC}}$ are the bit-costs produced by the chosen hybrid compression methods, i.e., the generative autoencoder and classical lossless compressor, respectively. $\mathbbm{1}_{\{\vm_s^T\vm_c >0\}}$ denotes the indicator function that returns 1 if at least one entity $i$ in the latent space is sampled and compressed using generative models  ($\vm_s[i]\vm_c[i]=1$), and 0 otherwise. This formulation ensures that, whenever any element associated with a latent dimension is selected for generative compression, the entire latent representation $\mX$ is transmitted. The constrained optimization problem is then formulated as:
\begin{align}
\label{eq:probForm}
\max_{f, g, \pi_s, \pi_c} & \EE_{\vc, \vm_s, \vm_c} \left[I(\mZ; \mY|\vc, \vm_s, \vm_c)\right]\\
\mbox{s.t. } & \eqref{eq:maskedCodec}, \eqref{eq:sampleCost}, \eqref{eq:rateCost}\nonumber\\
& \Sss(\pi_s)\leq S(\rho_s), \nonumber\\
& \Rs(\pi_s, \pi_c, f)\leq R(\rho_c), \nonumber
\end{align}
where $S(\rho_s)$ and $R(\rho_c)$ are the sampling budget and rate budget reflecting the memory size $\rho_s$ at the transmitter and the communication bandwidth $\rho_c$, respectively.

\section{Proposed Solution}
\label{sec:proposed_solution}

To solve the problem described in the previous section, we propose a Goal-Oriented Generative Hybrid Sampling and Compression (GO-GenZip) scheme, whose architecture is depicted in Fig.~\ref{fig:architecture}. Our solution implements the following steps to achieve data reduction and compression: (1) first, we sample the original data $\mb{X}$ generated at the source according to a mask $\vm_s$; (2) the sampled data $\vm_s \odot \mb{X}$ are then divided into two subsets to leverage the hybrid compression schemes, as defined by $\vm_c$; (3) the two data segments are compressed separately, one through a lossless scheme $E_{\operator{LC}}$ such as the compressor \gls{lzma} and the other through our developed generative encoder model, $E_{\operator{GE}}$. Therefore, the representation of compressed data consists of the combination of the latent representation of the GenAI model and compressed samples at the lossless compressor. The compressed data are transferred over the network to reach a centralized server. Finally, data is reconstructed via the lossless decoder, $D_{\operator{LC}}$, and the generative decoder model, $D_{\operator{GE}}$. The original data are restored by merging the data coming from the two compression methods. In the following subsections, we detail the formulation of the sampling policy and of the hybrid compression scheme. Optionally, a task (App $i$) that estimates the target $\mb{\hat{Y}}$ can be considered. It is defined as a function of the reconstructed data, which means that for \gls{go} training, the application block is added and the model is optimized end-to-end. 

\subsection{Sampling Policy}
In the adaptive setting, the policy $\pi_s$ is based on the context information $\mb{c}$. A fully connected neural network uses this information to obtain log-probabilities (logits) of the sampling and hybrid compression policy. During training, the \emph{Gumbel-Softmax distribution} \cite{jang2016categorical} is employed to enable differentiable sampling from a categorical distribution, which facilitates learning a stochastic policy. Given a categorical distribution with class probabilities $\mb{p} = (p_1, p_2, \ldots, p_k)$, the Gumbel-Softmax sample $\mb{y} = (y_1, y_2, \ldots, y_k)$ is computed as
\begin{equation}
y_i = \frac{\exp\left((\log p_i + g_i)/\tau \right)}{\sum_{j=1}^k \exp\left((\log p_j + g_j)/\tau \right)},
\end{equation}
where $g_i$ are independent and identically distributed samples from the Gumbel(0,1) distribution, and $\tau > 0$ is the temperature parameter controlling the smoothness of the approximation. As $\tau \to 0$, the Gumbel-Softmax distribution approaches a one-hot vector, approximating a discrete sample, while for larger values of $\tau$, the output remains soft and differentiable. This property allows the policy $\pi_\theta$ parameterized by $\theta$ to be updated via gradient-based optimization methods despite the inherently discrete nature of action sampling. 

To train the policy, \emph{\gls{st} gradient estimation} \cite{liu2023bridging} is used, leading to an effective update of the log-probabilities. Similarly to the reparameterization trick in \gls{vae}, \gls{st} allows us to model the output of the policy as a discrete choice in the forward pass, while only the choice probability (soft choice) is considered in the backward pass. Let $\mb{X}_s$ be the discrete sampled choices for the policy $\pi_\theta(\mb{c})$; the reparameterized vector $\tilde{\mb{X}}_s$ is
\begin{equation}
    \tilde{\mb{X}}_s = \mb{y} \odot \mb{X}_s + \text{sg}((1-\mb{y}) \odot \mb{X}_s),
\end{equation}
where $\text{sg}(\cdot)$ is the stop gradient operation which detaches the input from the gradient tracking.

\subsection{Hybrid Compression}
Similarly to the sampling policy, a compression policy $\pi_c$ is used to obtain the mask $\vm_c$ in which some of the acquired entries are set to zero, meaning that these samples will be compressed by lossless compression based on entropy. Specifically, this exploits the reparameterization trick already explained for the sampling policy. The masked values in this case are compressed with the standard \gls{lzma} compression algorithm and converted into a string of bits.
Since the sampling mask and the hybrid compression mask are binary, the data selected for the \gls{genai} compression method are obtained as $\mb{m}_c \odot \mb{m}_s \odot \mb{X}$. 
As a result, the compressed sample $\mb{Z}$ is the combination of two formats: the latent representation of the autoencoder and the string of bytes generated by the \gls{lzma} algorithm. 

\subsection{Dual Optimization and Constraint Matching}
Our solution jointly optimizes $\pi_s$ and $\pi_c$ by representing them as a $D\times3$ matrix $\mb{M}\in\mathbb{R}^{D\times3}$: each row $d$ of the matrix has $3$ columns, representing the log-probability of not sampling the entry $d$, sampling and compressing it with the GenAI model, and sampling and compressing with \gls{lzma}, respectively. The Gumbel-Softmax approach is then used for joint sampling and hybrid compression, improving target accuracy.
To include the sampling and compression constraints $S(\rho_s)$ and $R(\rho_c)$ in the loss function, we introduce the dual parameters $\beta_s$ and $\beta_c$. When we consider pure data reconstruction ($\mb{Y} = \mb{X}$), the loss function can be written as the Lagrangian of the original loss, using multipliers $\beta_s$ and $\beta_c$ to weight the constraint functions:
\begin{equation}
\label{eq:loss}
    \mc{L} \!=\! ||\mb{X} - \mb{\hat{X}}||^2_2 + \beta_s (\mc{S}(\pi_s) - S(\rho_s)) + \beta_c (\mc{R}(\pi_s, \pi_c, f)-R(\rho_c)),
\end{equation}
where the terms $\beta_s, \,\mc{S}(\pi_s) $ and $\beta_c, \,\mc{R}(\pi_s, \pi_c, f)$ are introduced into the loss function to force the policy to match the constraints. For each combination of $\beta_s$ and $\beta_c$, the optimization procedure converges to specific sampling and compression rates. Target sampling and compression rates can be obtained by adjusting the dual parameters until the constraints are met. 

\begin{figure}[t]
\vspace{-8pt}
\centering
\begin{algorithm}[H]
\caption{Training Loop of Go-GenZip}
\label{alg:gogenzip}
\begin{algorithmic}[1]
\footnotesize
  \Require $E_{gen}, D_{gen}, \pi_s, \pi_c$ parametrized by $\theta$. $\beta_s \gets 0$, $\beta_c \gets 0$. $\lambda$ (learning rate).
  \For{$B \in \mc{D}$}
    \State $\mb{m}_s \gets \pi_s(\mb{c}),\ \mb{m}_c \gets \pi_c(\mb{c})$
    \State $\mb{X}_g \gets \mb{m}_c \odot \mb{m}_s \odot \mb{X}$
    \State $\mb{X}_h \gets (1 - \mb{m}_c) \odot \mb{m}_s \odot \mb{X}$
    \State $\hat{\mb{X}} = D_{\text{GE}}(E_{\text{GE}}(\mb{X}_s)) + D_{\text{LC}}(E_{\text{LC}}(\mb{X}_h))$ 
    \State Update $\theta$ according to \eqref{eq:loss}
    \State Update $\beta_s$ and $\beta_c$ according to \eqref{eqn:upt_beta_s} and \eqref{eqn:upt_beta_c}, respectively 
     
  \EndFor
  
\end{algorithmic}
\end{algorithm}
\vspace{-0.8cm}
\end{figure}

Alg.~\ref{alg:gogenzip} shows the pseudocode for one iteration of the training procedure of \gls{gogenzip}. $\mc{D}$ represents the set of all training batches; for each batch, we update the autoencoder and the two policies according to the loss in \eqref{eq:loss}. The reconstructed data $\hat{\mb{X}}$ is then obtained as the combination of the reconstructed data from the two compression methods. It should be noted that the expected sampling rate and compression rate are estimated as the average over a batch to avoid computing the expectation over the entire training dataset. 

This training loop shows dual optimization of the unconstrained problem in~\eqref{eq:loss}, while the coefficients are updated as\footnote{$[x]_{+}=\max(0,x)$ ensures the non-negativity of the multipliers.}
\begin{align}
    &\beta_s \gets \left[\beta_s + \lambda (\Sss(\pi_s) - S(\rho_s))\right]_{+}, \label{eqn:upt_beta_s}\\
    &\beta_c \gets \left[\beta_c + \lambda (\Rs(\pi_s, \pi_c, f) -  R(\rho_c))\right]_{+}\label{eqn:upt_beta_c}
\end{align}
to enforce the optimization constraints.

\subsection{Goal-Oriented Training}
The reconstruction set is a special case of the \gls{ib} system model, in which the target $\mb{Y}$ corresponds to the input data $\mb{X}$. Here, we propose a more general training modality that is more suitable for \gls{go} sampling and compression. 
Since we consider multi-dimensional time-series as entry data, we focus on prediction tasks as optimization goals. Specifically, we collect $L$ lookback samples $\mb{X}_{t-L+1:t}:=\left[\mb{X}_{t-L+1} ,...,\mb{X}_t\right]\in\R^{K\times L}$ and use them as input data for a prediction module $g: \mathbb{R}^{K\times L} \to \mc{Y}$. For example, for a given prediction task, the target data $\hat{\mb{Y}}$ are obtained considering an horizon $H$, a target $k$-th \gls{kpi}, and a specific function:
$\phi\in\{ \text{identity, mean, min, max} \}$,
such that $\mb{Y} = \phi\left(\mb{X}^{(k)}_{t+1:t+H}\right)$, where $\mb{X}^{(k)}_{t+1:t+H}$ denotes the next $H$ samples of the $k$-th KPI.
This creates a wide range of prediction tasks, each requiring attention to different temporal or feature dependencies within the input data. The required level of accuracy on the reconstructed data varies with respect to the prediction function. The training of the model is similar to the data reconstruction task: a specific sampling rate and compression rate constrain the system model, while the the policy and the prediction module try to maximize task performance. However, in the multi-task scenario, a batch of training data is selected to contain different tasks in the same training step, and the context information $\mb{c}$ is extended to contain the task identifier. 
Adding the task embedding to the context information realizes the full \gls{go} strategy, enabling automatic discovery of the adaptive sampling and hybrid compression strategies. 

\section{Experimental Results}
In this section, we evaluate our model on real network telemetry data collected by $1162$ 4G \glspl{bs} deployed across different regions over a ten-day period, with an hourly sampling interval.  Each \gls{bs} reports 34 \glspl{kpi} at each hour, covering diverse network performance metrics, such as throughput-, latency-, and mobility-related metrics. The data are preprocessed to cluster \glspl{bs} with similar traffic patterns, enabling the extraction of contextual information including \gls{bs}-class labels. This contextual information supports the learning of specialized sampling and compression policies. The final context vector includes the embeddings of \emph{\gls{bs} class}, \emph{hour of the day}, and \emph{task identifier} (for the multi-task goal-oriented scheme). 

The system comprises a hybrid multi-task compression architecture with two main components: a \emph{MaskedModel} for adaptive sampling and compression and a \emph{MultiTaskModel} for multi-task prediction. The former uses an adaptive policy network (two-layer MLP with ELU activation) that maps contextual metadata to per-\gls{bs} compression decisions via Gumbel-Softmax sampling, selecting among no sampling, \gls{ae}-based generative compression, or \gls{lzma} compression. The \gls{ae} employs a conditional encoder-decoder with configurable latent dimensions determining compression ratios. The latter consists of task-specific predictors with layers (128,~64,~32) with GELU activation and dropout.

\subsection{Reconstruction Performance of GO-GenZip}
We conducted a first test on data reconstruction accuracy of \gls{gogenzip}, comparing its performance with lossless compression in terms of compression ratio. Also, we evaluated the impact of reducing the sampling fraction and constraining the system to limited observability. We tracked the \gls{maer} over multiple configurations of the proposed model. 

\begin{figure}
    \centering
    \begin{tikzpicture}
        \begin{axis}[
            width=8cm,
            height=5cm,
            grid=major,
            xlabel={Sampling Ratio (SR)},
            ylabel={Mean Absolute Error},
            ytick={0, 0.05, 0.1},
            yticklabels={0.0, 0.05, 0.1},
            legend style={at={(0.5,1.4)}, anchor=north, cells={align=left}, legend columns=3, font=\scriptsize},
        ]
        \addplot[
            color=mplblue,
            mark=square,
            dashed,
            mark options={solid},
            thick
        ] table [x index=0, y index=1, col sep=space] {figures/data_for_paper/CR_4.25_S-G.dat};
        \addlegendentry{CR 4.25 S-G}
        \addplot[
            color=mplorange,
            mark=o,
            dashed,
            mark options={solid},
            thick
        ] table [x index=0, y index=1, col sep=space] {figures/data_for_paper/CR_2.83_S-G.dat};
        \addlegendentry{CR 2.83 S-G}
        \addplot[
            color=mplgreen,
            mark=star,
            dashed,
            mark options={solid},
            thick
        ] table [x index=0, y index=1, col sep=space] {figures/data_for_paper/CR_2.12_S-G.dat};
        \addlegendentry{CR 2.12 S-G}
        \addplot[
            color=mplblue,
            mark=square,
            thick
        ] table [x index=0, y index=1, col sep=space] {figures/data_for_paper/CR_4.59_S-H.dat};
        \addlegendentry{CR 4.59 S-H}
        
        \addplot[
            color=mplorange,
            mark=o,
            thick
        ] table [x index=0, y index=1, col sep=space] {figures/data_for_paper/CR_2.93_S-H.dat};
        \addlegendentry{CR 2.93 S-H}
        
        \addplot[
            color=mplgreen,
            mark=star,
            thick
        ] table [x index=0, y index=1, col sep=space] {figures/data_for_paper/CR_2.11_S-H.dat};
        \addlegendentry{CR 2.11 S-H}
        \addplot[
            color=red,
            mark=x,
            thick
        ] coordinates {(1, 0)};
        \addlegendentry{lossless}
        \end{axis}
    \end{tikzpicture}
    \caption{Comparison between Generative compression and Hybrid compression for different SRs.}
    \label{fig:p_res_sr}
    \vspace{-0.5cm}
\end{figure}

\begin{figure}
    \centering
    \begin{tikzpicture}
        \begin{axis}[
            width=8cm,
            height=5cm,
            grid=major,
            xlabel={Compression Ratio (CR)},
            ylabel={Mean Absolute Error},
            ytick={0, 0.05, 0.1},
            yticklabels={0.0, 0.05, 0.1},
            legend style={at={(0.5,1.4)}, anchor=north, cells={align=left}, legend columns=3, font=\scriptsize},
        ]
        \addplot[
            color=mplblue,
            mark=square,
            dashed,
            mark options={solid},
            thick
        ] table [x index=0, y index=1, col sep=space] {figures/data_for_paper/SR_0.21_S-G.dat};
        \addlegendentry{SR 0.21 S-G}
        \addplot[
            color=mplorange,
            mark=o,
            dashed,
            mark options={solid},
            thick
        ] table [x index=0, y index=1, col sep=space] {figures/data_for_paper/SR_0.4_S-G.dat};
        \addlegendentry{SR 0.4 S-G}
        \addplot[
            color=mplgreen,
            mark=star,
            dashed,
            mark options={solid},
            thick
        ] table [x index=0, y index=1, col sep=space] {figures/data_for_paper/SR_1.0_S-G.dat};
        \addlegendentry{SR 1.0 S-G}
        \addplot[
            color=mplblue,
            mark=square,
            thick
        ] table [x index=0, y index=1, col sep=space] {figures/data_for_paper/SR_0.22_S-H.dat};
        \addlegendentry{SR 0.22 S-H}
        
        \addplot[
            color=mplorange,
            mark=o,
            thick
        ] table [x index=0, y index=1, col sep=space] {figures/data_for_paper/SR_0.4_S-H.dat};
        \addlegendentry{SR 0.4 S-H}
        
        \addplot[
            color=mplgreen,
            mark=star,
            thick
        ] table [x index=0, y index=1, col sep=space] {figures/data_for_paper/SR_1.0_S-H.dat};
        \addlegendentry{SR 1.0 S-H}
        \addplot[
            color=red,
            mark=x,
            thick
        ] coordinates {(1.37, 0)};
        \addlegendentry{lossless}
        \end{axis}
    \end{tikzpicture}
    \caption{Comparison between Generative compression and Hybrid compression for different CRs.}
    \label{fig:p_res_cr}
\end{figure}

\subsubsection{Comparison between hybrid and generative compression approaches} To assess the impact of hybrid compression, we trained the model under two configurations: (i) {\bf S–G}, which employs a sampling policy combined with a solely \gls{genai}-based compression module, and (ii) {\bf S–H}, which employs a sampling policy together with our proposed hybrid compression policy. Extensive results on the benefit of the hybrid compression scheme are presented in Fig.~\ref{fig:p_res_sr} and \ref{fig:p_res_cr}. We use the same color to highlight performance curves corresponding to the same \gls{cr} or \gls{sr}, respectively. These results confirm that the hybrid compression (\textbf{S-H}), in solid line, consistently improves performance. In Fig.~\ref{fig:p_res_sr} the {\bf S–H} curves are always below the corresponding {\bf S–G} plots, meaning lower reconstruction error for same CR. Similarly, in Fig.~\ref{fig:p_res_cr} the {\bf S–H} scheme obtains better performance when compared to {\bf S–G}. Only when the SR is very low ($\approx0.2$) the performance of the two methods are equivalent. This is confirmed also from Fig.~\ref{fig:p_res_sr} where all the models obtains similar MAEs regardless of the compression method and the compression ratio. This thorough testing approach demonstrates that the hybrid method not only performs well under various conditions but also maintains its accuracy advantage regardless of the specific compression or sampling settings, highlighting its robustness and reliability. 

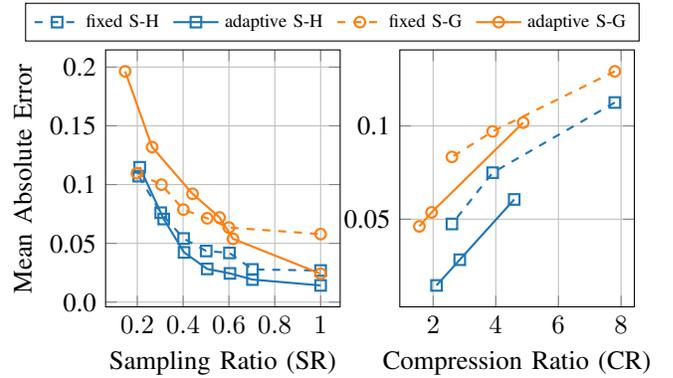
\begin{figure}
    \centering
    \subfloat{
\begin{tikzpicture}

\begin{axis}[
    width=0cm,
    height=0cm,
    axis line style={draw=none},
    tick style={draw=none},
    at={(0,0)},
    scale only axis,
    xmin=0,
    xmax=0,
    xtick={},
    ymin=0,
    ymax=0,
    ytick={},
    axis background/.style={fill=white},
    legend style={legend cell align=left, fill opacity=1, align=center, draw=black, font=\scriptsize, at={(0, 0)}, anchor=center, /tikz/every even column/.append style={column sep=0.3em}},
    legend columns=4,
]
\addplot [thick, mplblue, const plot mark left, dashed, mark=square, mark options={solid}]
table {%
0 0
};
\addlegendentry{fixed S-H}
\addplot [thick, mplblue, const plot mark left, mark=square]
table {%
0 0
};
\addlegendentry{adaptive S-H}
\addplot [thick, mplorange, const plot mark left, dashed, mark=o, mark options={solid}]
table {%
0 0
};
\addlegendentry{fixed S-G}
\addplot [thick, mplorange, const plot mark left, mark=o]
table {%
0 0
};
\addlegendentry{adaptive S-G}

\end{axis}

\end{tikzpicture}} \\ 
    \vspace{0.1cm}
    \setcounter{subfigure}{0}
    \begin{tikzpicture}
        \begin{axis}[
            width=4.7cm,
            height=5cm,
            grid=major,
            xlabel={Sampling Ratio (SR)},
            ylabel={Mean Absolute Error},
            ytick={0, 0.05, 0.1, 0.15, 0.2},
            yticklabels={0.0, 0.05, 0.1, 0.15, 0.2},
        ]
        \addplot[
            color=mplblue,
            mark=square,
            mark options={solid},
            dashed,
            thick
        ] table [x index=0, y index=1, col sep=space] {figures/data_for_paper/adaptive_S-G_sr.dat};
        \addplot[
            color=mplblue,
            mark=square,
            thick
        ] table [x index=0, y index=1, col sep=space] {figures/data_for_paper/adaptive_S-H_sr.dat};
        \addplot[
            color=mplorange,
            mark=o,
            dashed,
            mark options={solid},
            thick
        ] table [x index=0, y index=1, col sep=space] {figures/data_for_paper/fixed_S-G_sr.dat};
        \addplot[
            color=mplorange,
            mark=o,
            mark options={solid},
            thick
        ] table [x index=0, y index=1, col sep=space] {figures/data_for_paper/fixed_S-H_sr.dat};
        \end{axis}
    \end{tikzpicture}
    \hspace{-0.5cm}
    \begin{tikzpicture}
        \begin{axis}[
            width=4.7cm,
            height=5cm,
            grid=major,
            xlabel={Compression Ratio (CR)},
            ytick={0, 0.05, 0.1, 0.15, 0.2},
            yticklabels={0.0, 0.05, 0.1, 0.15, 0.2},
        ]
        \addplot[
            color=mplblue,
            mark=square,
            thick,
            dashed,
            mark options={solid}
        ] table [x index=0, y index=1, col sep=space] {figures/data_for_paper/adaptive_S-G.dat};
        \addplot[
            color=mplblue,
            mark=square,
            thick
        ] table [x index=0, y index=1, col sep=space] {figures/data_for_paper/adaptive_S-H.dat};
        \addplot[
            color=mplorange,
            mark=o,
            thick,
            dashed,
            mark options={solid}
        ] table [x index=0, y index=1, col sep=space] {figures/data_for_paper/fixed_S-G.dat};
        \addplot[
            color=mplorange,
            mark=o,
            thick
        ] table [x index=0, y index=1, col sep=space] {figures/data_for_paper/fixed_S-H.dat};
        \end{axis}
    \end{tikzpicture}
    \caption{Comparison of adaptive and fixed method with respect to SR and CR.}
    \label{fig:comparison}
    \vspace{-0.5cm}
\end{figure}


\begin{figure}
    \centering
    \begin{tikzpicture}
\begin{axis}[
    width=4.5cm,
    height=3cm,
    axis on top,
    enlargelimits=false,
    xmin=0, xmax=34,
    ymin=0, ymax=24,
    point meta min=0,
    point meta max=1,
    xlabel = KPI Index,
    ylabel = Hour,
    title = Task 3: Latency - BS 0,
]
\addplot graphics [xmin=0, xmax=34, ymin=0, ymax=24] {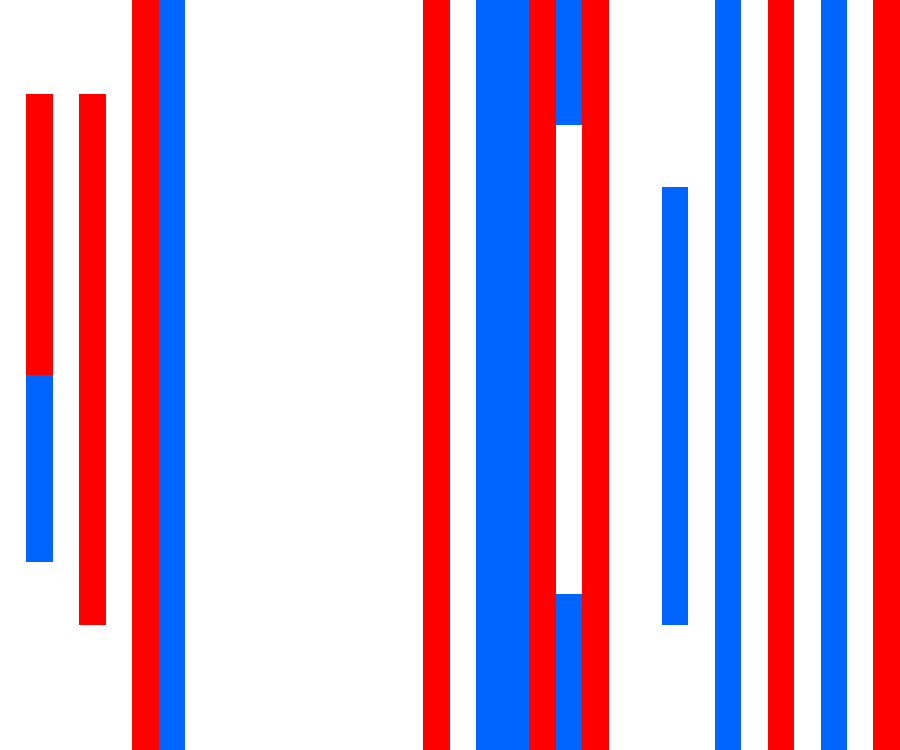};

\end{axis}
\end{tikzpicture}
    \begin{tikzpicture}
\begin{axis}[
    width=4.5cm,
    height=3cm,
    axis on top,
    enlargelimits=false,
    xmin=0, xmax=34,
    ymin=0, ymax=24,
    point meta min=0,
    point meta max=1,
    xlabel = KPI Index,
    title = Task 3: Latency - BS 1,
]
\addplot graphics [xmin=0, xmax=34, ymin=0, ymax=24] {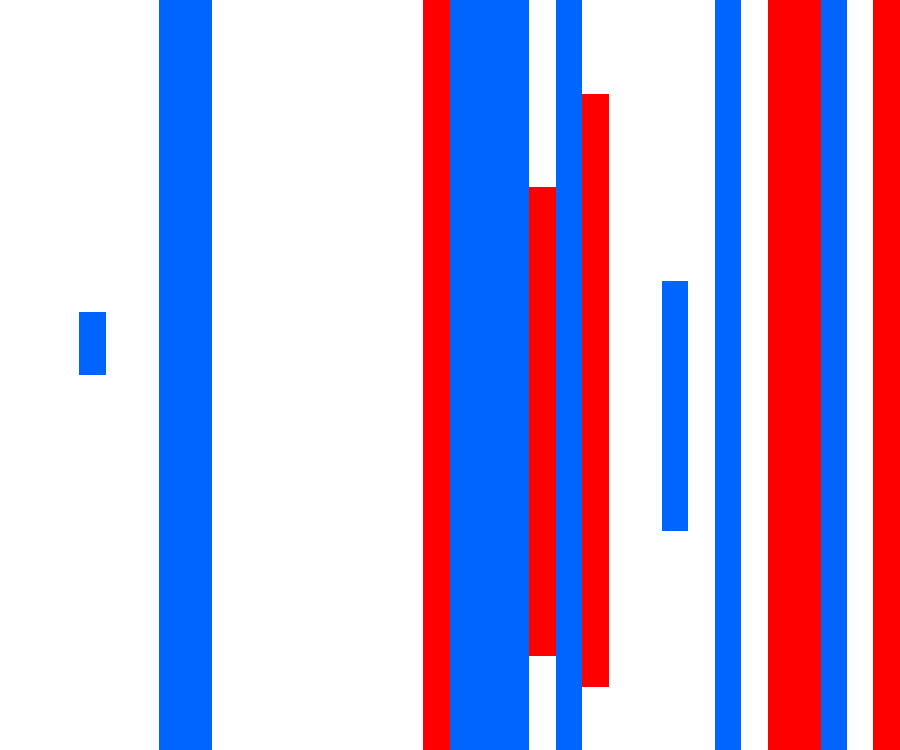};

\end{axis}
\end{tikzpicture}\\
        \begin{tikzpicture}
\begin{axis}[
    width=4.5cm,
    height=3cm,
    axis on top,
    enlargelimits=false,
    xmin=0, xmax=34,
    ymin=0, ymax=24,
    point meta min=0,
    point meta max=1,
    xlabel = KPI Index,
    ylabel = Hour,
    title = Task 4: UL Rate - BS 0,
]
\addplot graphics [xmin=0, xmax=34, ymin=0, ymax=24] {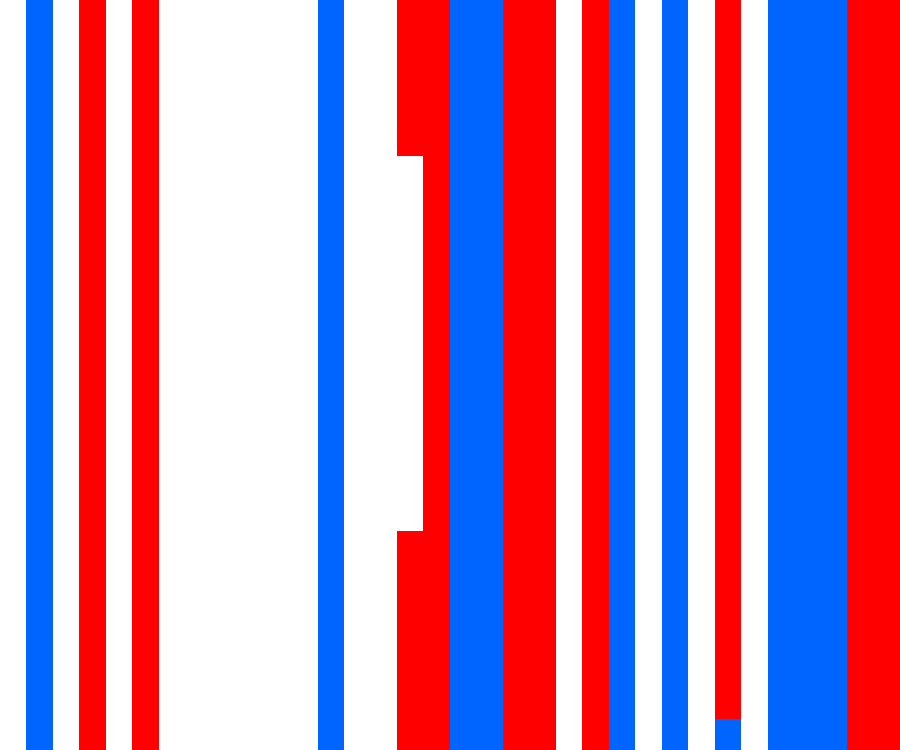};

\end{axis}
\end{tikzpicture}
    \begin{tikzpicture}
\begin{axis}[
    width=4.5cm,
    height=3cm,
    axis on top,
    enlargelimits=false,
    xmin=0, xmax=34,
    ymin=0, ymax=24,
    point meta min=0,
    point meta max=1,
    xlabel = KPI Index,
    title = Task 4: UL Rate - BS 1,
]
\addplot graphics [xmin=0, xmax=34, ymin=0, ymax=24] {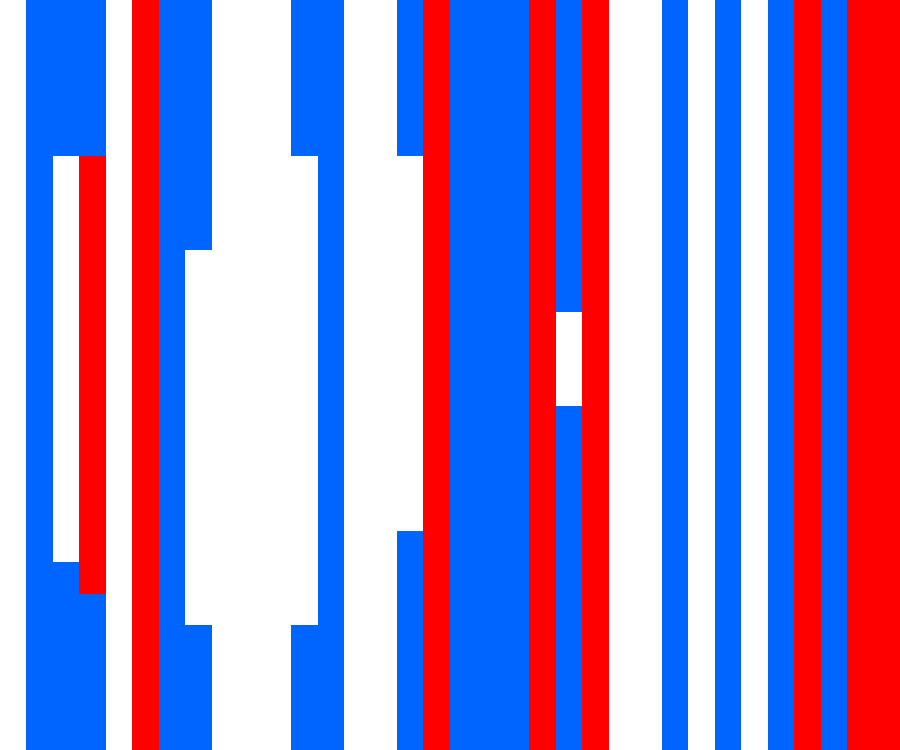};

\end{axis}
\end{tikzpicture}
    \caption{Visualization of masks for different BS classes and prediction tasks. White, blue, red regions represent the unsampled, compression with generative model, and compression with \gls{lzma}, respectively.}
    \label{fig:mask_visualization}
    \vspace{-3ex}
\end{figure}

\subsubsection{Comparison between adaptive and fixed policies} 
The adaptive strategy is compared with a fixed compression and sampling policy. In the fixed case, one sampling mask is learned for all the possible input $\mb{X}$, regardless of the conditional information. 
Fig.~\ref{fig:comparison} clearly illustrates that the adaptive solution significantly outperforms the fixed policy when evaluating the performance with respect to the \gls{sr}. Specifically, the adaptive approach adjusts dynamically to the context conditions, resulting in a more efficient allocation of sampling resources. This adaptability allows it to maintain a higher sampling ratio where needed, improving overall system performance and responsiveness. In contrast, the fixed policy, with its static nature, fails to optimize sampling in varying scenarios, leading to less effective outcomes. The two plots in Fig.~\ref{fig:comparison} show that the adaptive policy improves both with respect to varying SR and CR. Together, these results emphasize the clear advantage of employing adaptive methods over fixed policies for better sampling efficiency and system effectiveness.

\subsection{Multi-Task Training}


We conducted Multi-task training for a dataset of 6 prediction tasks using $L=3$ and $H=3$ for six \glspl{kpi}: \emph{downlink physical resource block usage}, \emph{radio resource connection}, \emph{latency}, \emph{downlink payload}, \emph{uplink rate}, \emph{handover attempt}. Due to the limited space, in Fig.~\ref{fig:mask_visualization} we report an example of obtained masks (white: unsampled, blue: generative compression, red: \gls{lzma} compression) for two selected prediction tasks, latency and uplink rate, and two BS classes, with target sampling ratio and compression ratio $(0.5, 0.5)$. For each of the plots, on the x-axis we report the \gls{kpi} index, while on the y-axis the hour of the day. Although certain structural similarities can be observed, distinct sampling patterns emerge across different tasks, confirming that the learned policies selectively emphasize \glspl{kpi} most relevant to each objective. Moreover, variations between \gls{bs} classes for the same task suggest that heterogeneous traffic conditions benefit from tailored sampling strategies. In addition, Table~\ref{tab:mae_comparison} presents the \gls{maer} for four tasks out of the six evaluated (due to the limited space) comparing models with and without GO end-to-end training. The goal-oriented  training consistently achieves lower errors and smaller variances, demonstrating enhanced robustness and stability across tasks.
\begin{table}[t]
\centering
\caption{Prediction \gls{maer} with and without GO end-to-end training.}
\setlength{\tabcolsep}{3pt}
\begin{tabular}{lcccccc}
\hline
\textbf{Method} & \textbf{Task1} & \textbf{Task2} & \textbf{Task3} & \textbf{Task4}  \\
\hline\hline
Recon-Based & 1.27±0.04 & 0.09±0.004 &  0.04±0.003 & 1.48±0.14 \\
GO E2E & 1.21±0.02 & 0.08±0.003 & 0.04±0.0008 & 1.18±0.07  \\
\hline
\end{tabular}
\vspace{-3ex}
\label{tab:mae_comparison}
\end{table}

\section{Conclusions}

The \gls{gogenzip} scheme presented in this paper proposes novel techniques to adaptively compress networking telemetry data based on the goal of the corresponding downstream tasks. This scheme adeptly manages diverse data characteristics, ranging from dense to sparse, by combining lossy generative compression techniques with lossless entropy-based compression. The masking policy has been proved to further augment the compression ratio, especially when task information is given as contextual information. The adopted scheme has shown promising results in both reconstruction accuracy and task objectives, providing high flexibility to adapt to different incoming network requests. Future research endeavors will focus on extending the proposed solution through the evaluation of various training schemes and validate the approach's efficacy across a broader spectrum of networking data types.

\bibliographystyle{IEEEtran}
\bibliography{biblio} 

\end{document}